\documentclass[runningheads]{llncs}
\usepackage{tikz}
\usepackage{pgfplots}
\usepackage{pgf-pie}
\usepackage{caption}
\usepackage{subcaption}
\usepackage{cite}
\usepackage{amsmath,amssymb,amsfonts}
\usepackage{algorithmic}
\usepackage{graphicx}
\usepackage{textcomp}
\usepackage{xcolor}
\usepackage{graphicx}
\usepackage{fancyhdr}
\usepackage{listings} 
\usepackage{changepage} 

\usepackage[colorinlistoftodos,prependcaption,textsize=tiny]{todonotes}

\newcommand{\accSex}{94.7}
\newcommand{\accAge}{68.0}
\newcommand{\accRace}{83.1}
\newcommand{\rmseAge}{8.33}
\newcommand{\maeAge}{6.21}

\newcommand{\pviMask}{0.853}
\newcommand{\pviNoMask}{0.828}
\newcommand{\diffPvi}{2.9}

\newcommand{\totalImages}{20,003}

\begin{document}
\title{Does a Face Mask Protect my Privacy?: Deep Learning to Predict Protected Attributes from Masked Face Images}
\titlerunning{Does a Face Mask Protect my Privacy?}
%
 \author{Sachith Seneviratne\inst{1} \and
 Nuran Kasthuriarachchi\inst{2} \and
 Sanka Rasnayaka\inst{4} \and
  Danula Hettiachchi\inst{3} \and Ridwan Shariffdeen\inst{4}}

 \authorrunning{S. Seneviratne et al.}


 \institute{University of Melbourne, Australia - 
 \email{sachith.seneviratne@unimelb.edu.au} \and
 University of Moratuwa, Sri Lanka -
 \email{nuran.11@cse.mrt.ac.lk}\\
 \and
 RMIT University, Australia -
 \email{danula.hettiachchi@rmit.edu.au} \and
 National University of Singapore -
 \email{\{sanka,ridwan\}@u.nus.edu}}

%
\maketitle              

\begin{abstract}
Contactless and efficient systems are implemented rapidly to advocate preventive methods in the fight against the COVID-19 pandemic. Despite the positive benefits of such systems, there is potential for exploitation by invading user privacy. In this work, we analyse the privacy invasiveness of face biometric systems by predicting privacy-sensitive soft-biometrics using masked face images. We train and apply a CNN based on the ResNet-50 architecture with \totalImages ~synthetic masked images and measure the privacy invasiveness. Despite the popular belief of the privacy benefits of wearing a mask among people, we show that there is no significant difference to privacy invasiveness when a mask is worn. In our experiments we were able to accurately predict sex (\accSex\%), race (\accRace\%) and age (MAE \maeAge ~and RMSE \rmseAge) from masked face images. Our proposed approach can serve as a baseline utility to evaluate the privacy-invasiveness of artificial intelligence systems that make use of privacy-sensitive information. We open-source all contributions for reproducibility and broader use by the research community.
\end{abstract}


\section{Introduction}

Since the outbreak of SARS-CoV-2 (COVID-19), the use of face masks has become ubiquitous around the world and has been identified as an important public health response to fight against the ongoing pandemic. The mass shift to wearing masks during the COVID-19 pandemic has radically changed the way in which many of our mundane activities are carried out. This situation demands the enablement of contactless and efficient operations, especially in retail services. Contactless technologies like face and iris based detection systems are pushed to reach newer heights, in contrast applications that rely on fingerprint recognition modalities suffer a significant loss due to the emerging requirements as an after-effect of the COVID-19 pandemic\cite{CARLAW20208}. In particular, face recognition is praised as one of the efficient and contactless means of verifying identity and prior research has studied the impact and techniques to improve face-recognition systems to further advance contactless operations\cite{Hariri2021EfficientPandemic,Cabani2021MaskedFace-NetCOVID-19}. Using computer vision to enhance contactless and efficient operations has shown promise in various applications (i.e. public compliance monitoring\cite{afkaar20}). In this work, we investigate the impact of using computer vision, specifically in face authentication systems for contactless identification and the possible implications on privacy. 
Despite the scalable automation it provides, face-recognition technology needs to adhere to the privacy regulations such as the General Data Protection Regulation (GDPR) and improve the perception of users to increase trust and acceptance. 

Considering the advancements in surveillance and monitoring technologies in response to COVID-19, the norms of acceptable information flow may shift. For instance, users' perspectives on the use of location information (which is privacy-sensitive information), has drastically changed in times of crisis\cite{Om21}. However, such temporary measures during a crisis may not prevail as a permanent and long-term acceptance because it would unnecessarily reduce a persons privacy. Although a wider acceptance of surveillance systems can be seen in the current situation, we argue that a popular misconception of, ``wearing face masks will increase privacy protection'' exists among most people. Therefore, we first investigate the perception of users with respect to face biometric solutions and their understanding of privacy protection. We conduct an online survey to extract the opinions of users on their privacy with face masks and we learn that generally users have higher confidence of privacy protection when using a face mask. In our study, we find that perceived privacy of wearing a mask is higher with a statistical significance (P=0.00964 $<$ 0.05).

Systems that use face biometrics could potentially reveal privacy-sensitive information such as soft-biometrics, which includes but are not limited to age, sex and race. Many of the artificial intelligence systems use such privacy-sensitive information but are restricted for the intended purposes. 
We evaluate the possible violations of privacy-protection in such systems that use face biometrics, with the use of masked face images and quantify the privacy invasiveness of such implementations. We implement several techniques to predict privacy-sensitive soft-biometrics such as age, sex and race, and we were able to achieve an accuracy of \accSex\% and \accRace\% in correctly classifying the sex and race, respectively. We were also able to accurately predict the age with an RMSE score of \rmseAge ~and MAE score of \maeAge. We then analysed the privacy invasiveness in our implementation for images with mask and without mask, to understand the privacy preservation when using a face mask. We show that there is no significant difference in privacy protection by quantifying the privacy invasiveness using the Privacy Vulnerability Index (PVI)\cite{9304922} for both settings, which recorded \textbf{only a \diffPvi\% difference} that implies no significance in wearing a mask. 


In this work, we make three contributions: \textbf{(1)} Quantitative analysis on privacy invasion on masked face images. To the best of our knowledge, we are the first to study the predictability of age, sex and race using masked face images. \textbf{(2)} Study the perception of privacy protection on wearing a face mask. Our results indicate that people consider masked faces to be less privacy invasive. In support of reproducible research, we open-source our model weights and scripts for the benefit of the research community. These models will enable future study on masked face biometric systems related to privacy protection. 

    
    

\section{Related Work}

\subsection{Biometrics and Privacy}
The use of biometrics have raised various privacy concerns due to the possibility of predicting protected attributes. Many studies have evaluated the predictability of soft-biometric attributes such as age, gender and race using common biometrics such as face\cite{guo2008locally}, iris\cite{thomas2007learning}, fingerprint\cite{badawi2006fingerprint}, voice \cite{childers1991gender} and gait\cite{9304922}. In this work, we go beyond than prediction and provide means of quantifying the privacy invasiveness in systems that use soft-biometric.

\subsection{Face Biometric and Masks} While computer vision research has examined face recognition methods robust to partial occlusions~\cite{Kim2005EffectiveOcclusion,Weng2016RobustRecognition}, with increased global mask use due to COVID-19, there is a renewed interest in masked face recognition. Recent work shows that current state of the art face recognition methods trained with full face images fail in accurately recognising masked faces \cite{Damer2020TheStudy}. Although researchers have created real-world masked face datasets ~\cite{Wang2020MaskedApplication,Cabani2021MaskedFace-NetCOVID-19}, there is limited work on developing specific machine learning models trained with masked images. In addition, face masks have also introduced a family of computer vision challenges such as mask detection~\cite{Loey2021APandemic}. While some prior work has implications on masked biometric analysis using masks\cite{9211002}, they have not used masked facial images for analysis. In particular, while biometric analysis focused around the ocular region can provide useful insights into masked facial analysis, we argue that only actual analysis on masked facial imagery provides realistic insights into masked biometric analysis. This arises from the fact that, based on the masking process used, considerable portions of the ocular region may be occluded as well. Therefore, performing end-to-end evaluation of masked facial images provides a more realistic picture of the situation corresponding to real world usage.

\subsection{User Perception} User perceptions towards biometric modalities tend evolve with time \cite{8698599}. State of the art face recognition methods can achieve high levels of accuracy and are widely used for different applications, including authentication and surveillance. While users are generally more familiar and comfortable with face biometric solutions~\cite{Buckley2019ThePerceptions}, users also tend to resist face recognition based solutions due to privacy concerns~\cite{Liu2021ResistanceFactors}. Furthermore, wearing masks can limit the face area exposed to face recognition systems. However, there is limited literature on how people perceive the difference between masked and non-masked face recognition. This understanding could be potentially influenced by other challenges humans face when wearing masks. For instance, research shows that people find it challenging to match familiar faces, match unfamiliar faces and recognise emotions when faces are occluded with objects such as masks and sunglasses~\cite{Noyes2021TheObservers}. 


%

\section{Methodology}
Our evaluation consists of three components. First, we conducted a survey to understand the perception of users on privacy protection while wearing  a mask, from the increased surveillance systems due to COVID-19. Second, we generate a synthetic face mask dataset, predict protected attributes from masked face images and compare our results with prior work that use non-masked faces. Third, we show how unmasked face images invade privacy and analyse the impact of image attributes on our predictions.

\subsection{User Perception Survey}

The main objective of this survey was to study the perceptions of people towards the privacy invasiveness of masked faces in comparison to unmasked face images. We aim to answer the following questions: \textit{``Do people feel that wearing a face mask will protect their privacy?''}
and \textit{``Which is considered more private among Age, Race and Sex''}. An online survey was designed to collect this information with \textit{Yes/No questions} comparing the privacy invasiveness of masked and unmasked face images, \textit{Three point Likert scale questions} evaluating perceived privacy invasiveness of masked images and unmasked images and a \textit{Sorting Activity} to sort Age, Race and Sex based on importance. The survey is conducted anonymously on a voluntary basis in June 2021. The relative ordering of the sorting activity will be used to measure the Relative Importance Index (RII) value for each of the three attributes. 

\subsection{Dataset and Synthetic Mask Generation}
\label{datasets}
\begin{figure*}
\centering
\begin{subfigure}{.2\textwidth}
  \centering
  \includegraphics[height=\linewidth]{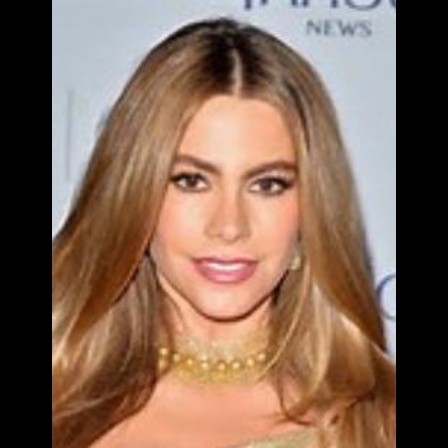}
  \caption{{Original}}
  \label{fig:sfig7}
\end{subfigure}
\begin{subfigure}{.2\textwidth}
  \centering
  \includegraphics[height=\linewidth]{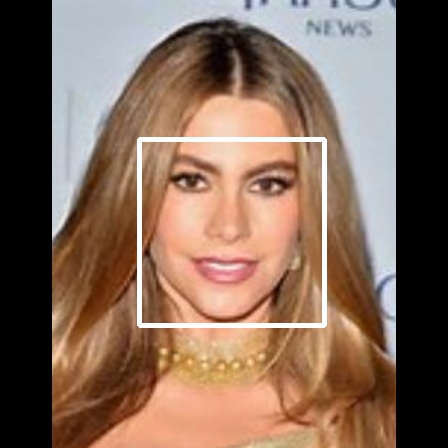}
  \caption{{Localization}}
  \label{fig:sfig8}
\end{subfigure}
\begin{subfigure}{.2\textwidth}
  \centering
  \includegraphics[height=\linewidth]{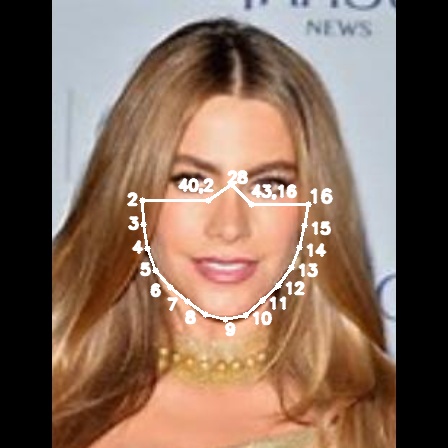}
  \caption{{Key points}}
  \label{fig:sfig9}
\end{subfigure}
\begin{subfigure}{.2\textwidth}
  \centering
  \includegraphics[height=\linewidth]{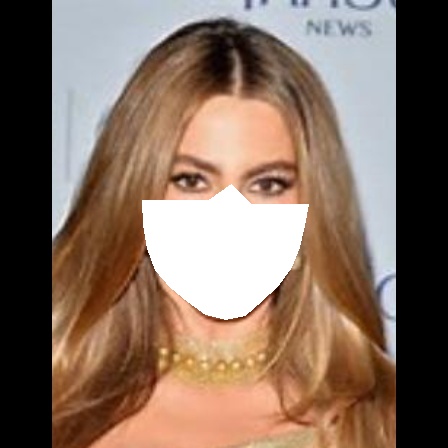}
  \caption{{Digital mask}}
  \label{fig:sfig10}
\end{subfigure}
\caption{\textbf{Synthetic mask creation pipeline (an example from UTK faces)}}
\label{fig:digimask}
\end{figure*}

There is no openly-available large-scale mask dataset with soft-biometric labels for age, gender and race. Therefore, we select UTK faces dataset, the most commonly cited face dataset in the literature and generate a masked dataset by digitally painting a mask on top of the face image.
We follow the process outlined in \cite{ngan2020ongoing} to generate synthetic masks on the face images. This process is depicted in Fig. \ref{fig:digimask}. We open-source the scripts used for  this process.


\begin{figure*}
\centering
\begin{subfigure}{.5\textwidth}
  \centering
  \includegraphics[width=1\linewidth]{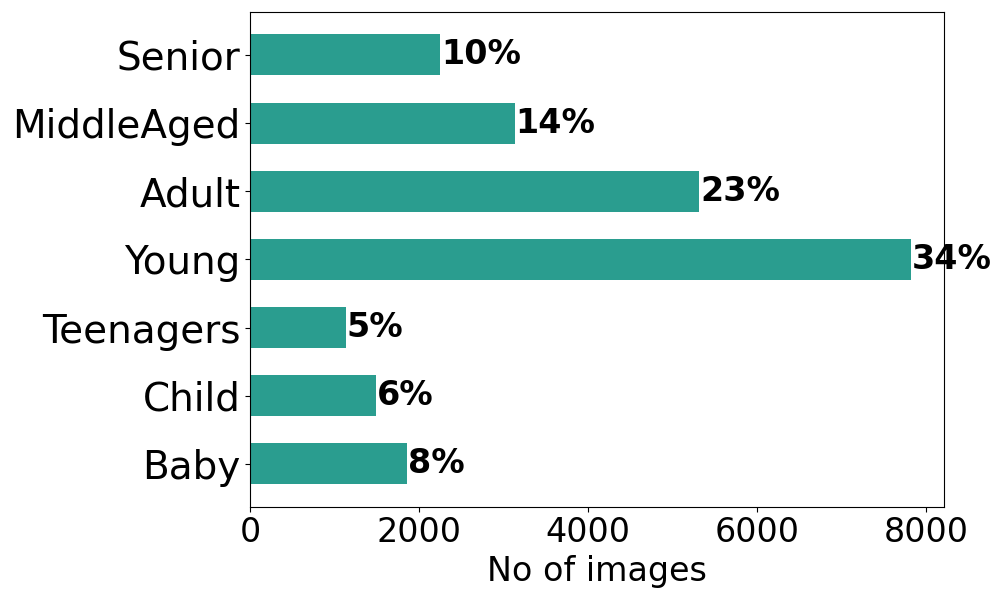}
  \caption{\textbf{Age distribution}}
  \label{fig:sfig5}
\end{subfigure}%
\begin{subfigure}{.5\textwidth}
  \centering
  \includegraphics[width=1\linewidth]{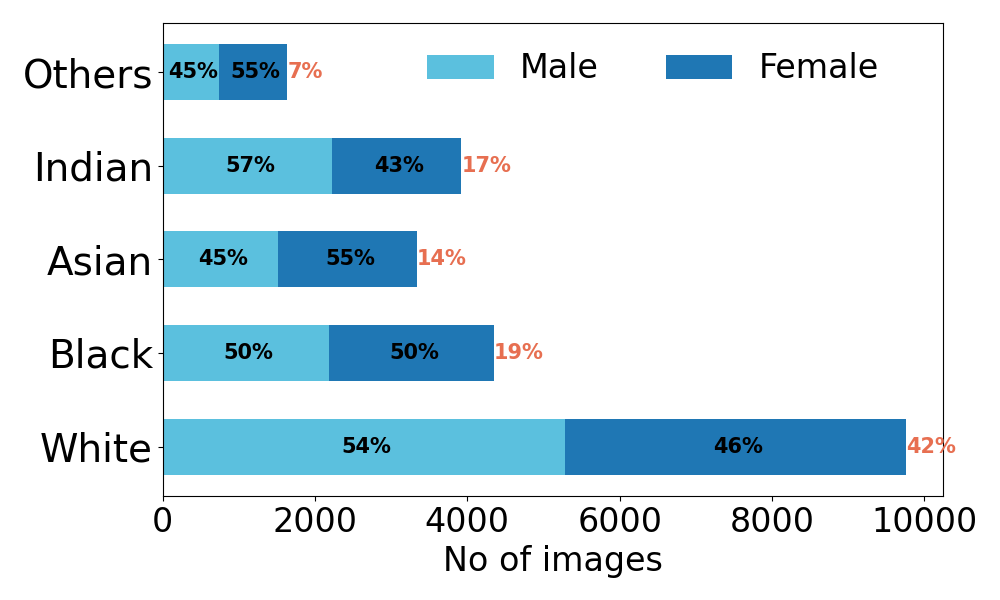}	  
  \caption{\textbf{Race and gender distribution}}
  \label{fig:sfig6}
\end{subfigure}
\caption{\textbf{UTK faces dataset summary}}
\label{fig:utksummary}
\end{figure*}

UTK faces dataset has 23,542 face images with labels for age, gender and race. Following the masking process\cite{ngan2020ongoing} we create a data-set of 23,002 masked images. We show the distributions of the attributes in Figure \ref{fig:utksummary}. We bin the ages as follows, baby: 0-3 years, child: 4-12 years, teenagers: 13-19 years, young: 20-30 years, adult: 31-45 years, middle aged: 46-60 years and senior: 61 years and above, in line with the analysis in \cite{das2018mitigating}.

\subsection{Computer Vision Workflow}

We use a computer vision based method with convolutional neural networks (CNN) to build models for 3 different tasks - age, sex and race prediction. Rather than building individual CNNs from scratch for each task, we build an initial facial representation within the neural network by using only the UTK facial dataset. We use a ResNet50 architecture and pre-train our representation for 3038 epochs in an unsupervised manner using the framework introduced in \cite{he2020momentum} and the projection head and augmentations from \cite{chen2020simple} using the parameters and settings described in \cite{chen2020improved}. Pre-training is carried out on a 4 GPU node with batch size set to 128. This pre-trained representation is then fine-tuned end-to-end for each specific task with a new fully connected layer incorporated past the final bottleneck layer of the pre-trained ResNet50 architecure. The size of this final output layer depends on the task. For regression the layer has a single node - as the output is a single continuous variable, while for classification we incorporate a number of nodes equal the number of classes in the problem (for instance, 2 for sex, 5 for race and 7 for age - corresponding to categorical labels). Training is carried out for 3500 epochs for each task with Stochastic Gradient Descent and learning rate 1e-3, and we isolate the single checkpoint with the best validation performance to evaluate on a holdout dataset. We evaluate the impact of additional image augmentations using RandAugment\cite{cubuk2019randaugment} with default ImageNet parameters. We open-source all contributions\footnote{https://github.com/sachith500/MaskedFaceRepresentation}, including trained models on the different splits and dataset splits for full reproducibility.

We evaluate against an open-sourced masked-facial representation  (MUFM) \cite{seneviratne2021multidataset} released as part of the Masked Facial Recognition Competition 2021~\cite{boutros2021mfr}, which claims to be a generic masked-face representation adaptable to any task on masked faces. We run this evaluation for the task of sex classification on a random split of 70\% training 20\% validation and 10\% testing on the UTK dataset. Based on results, we extend our analysis using the best performing combination of representation and technique for the tasks of race classification and age regression on similar random splits of UTK dataset (see Table~\ref{tab:model_results}). As a follow up experiment we build models on a new split of UTK data that ensures a uniform split as discussed in Section~\ref{datasets}. We train models as before (discarding previously trained models), but change the age regression to an age bracket based classification following other work in the literature \cite{das2018mitigating}. By doing so we compare against multiple existing state of the art techniques for age prediction (see Table~\ref{tab:result_comparison}). Note that our models are at a disadvantage due to roughly half of the face being absent/occluded in the image.

\subsection{Privacy Vulnerability Index(PVI)}
The Privacy Vulnerability Index \cite{9304922} is used to quantify the privacy invasiveness of a biometric modality. We use this measure to compare the privacy invasiveness of face images and masked face images. The PVI of a biometric depends on two factors,  \textbf{(1) Predictability ($p_i$):} how well can protected attributes be predicted using the biometric modality, measured by classification accuracy. \textbf{(2) Importance ($s_i$) :} how important is each personal attribute, measured using the RII calculated from the user perception study. 
The PVI value for masked and unmasked images is calculated as a weighted sum of these two values using the equation, $PVI={(\sum\nolimits_{i} s_i*p_i)}/{\sum\nolimits_{i} s_i}$.



\section{Evaluation Results}

\subsection{User Perception Study}

\begin{figure}[htb]
\centering
\begin{subfigure}{0.8\textwidth}
  \centering
  \includegraphics[width=1\linewidth]{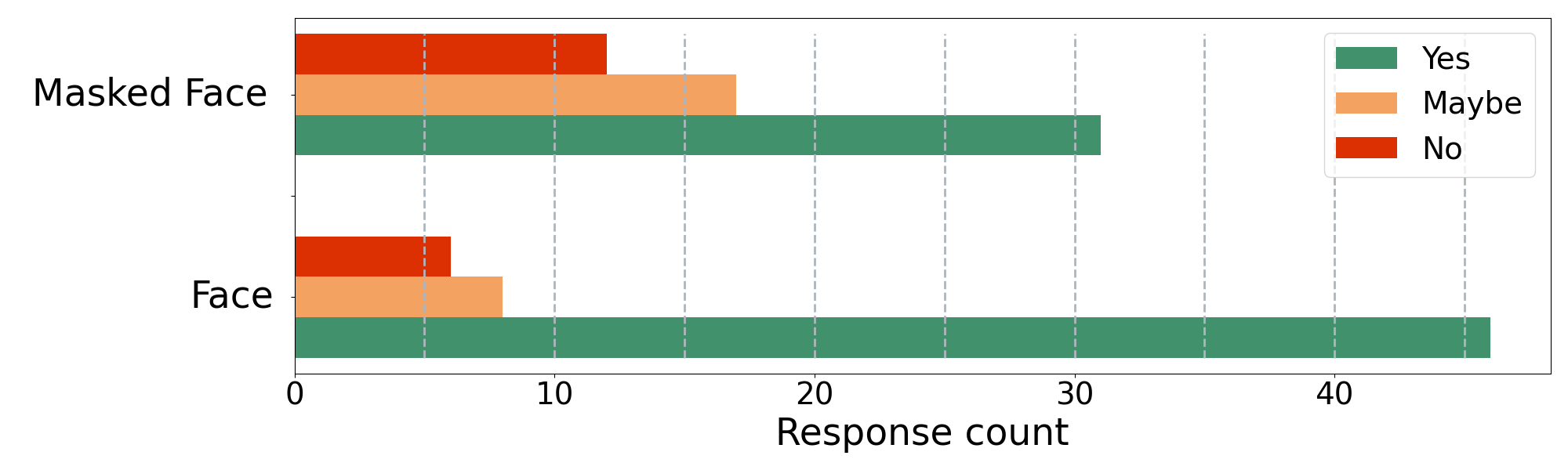}
  \label{fig:sfig1}
\end{subfigure}%
\caption{\textbf{User perception on privacy invasion with masked face images campared to face images.}}
\label{fig:utksummary2}
\end{figure}

The survey resulted in 60 complete responses. The users' responses to if the face image and masked face image could lead to privacy invasiveness is used to examine if there is a statistically significant difference in the perception towards the two modalities. We perform a the Mann-Whitney U test with a single-tail, to show that the perceived privacy of wearing a mask is higher with statistical significance ($P=0.00964 < 0.05$). Figure. \ref{fig:utksummary2} show the distribution of user responses.




Survey participants were asked if wearing a mask preserves privacy (compared to not wearing a mask). 50\% of the participants said yes while 40\% felt both violated privacy equally and 10\% said masked face images violates privacy more.

\textit{Relative Importance:} The resulting relative ordering with the Relative Importance Index (RII) values within brackets is; 1) Age [0.3765] 2)   Race [0.3353] 3) Gender [0.2882].

\subsection{Prediction Accuracy}

Table~\ref{tab:model_results} presents the overall accuracy for models built for masked face images. Table~\ref{tab:result_comparison} compares the results of masked and unmasked faces.

\begin{table}[ht]
\centering
\caption{\textbf{Attribute prediction using masked face images. The first experiment (sex) was used to verify that our model performance was superior to existing masked representations.}}
\label{tab:model_results}
\begin{tabular}{|l|l|l|l|l|} \hline
& Sex & Race & \multicolumn{2}{c|}{Age} \\
Method & Accuracy & Accuracy & MAE & RMSE \\
\hline
Using representation \cite{seneviratne2021multidataset} + transforms~\cite{cubuk2019randaugment} & 0.9374 &-&-&-\\
Our method with transforms from~\cite{cubuk2019randaugment} & \textbf{0.9401} & \textbf{0.8220} & 6.2788 & 8.4836\\
Our method without complex transforms & 0.9361 & 0.8134 & \textbf{6.2168} & \textbf{8.3372}\\
\hline
\end{tabular}
\end{table}















\begin{table}[htb]
\centering
\caption{\textbf{Overall result comparison with SOTA for each protected-attribute. Models are retrained for the uniform split using \textbf{optimal parameters} from experiments in Table~\ref{tab:model_results}.}}
\label{tab:result_comparison}
\begin{tabular}{|l|l|l|l|} \hline
                     &  & Masked Face & Masked Face\\ 
                     & Unmasked Face - SOTA & (Random Split) & (Uniform Split)\\ \hline
Sex                  & \textbf{\cite{das2018mitigating} 98.23\%} & 94.01\% &  \textbf{94.65\%}     \\ \hline
Race                 & \textbf{\cite{ahmed2020race} 91.23\%  }  &  82.20\% &  \textbf{83.12\%}       \\ \hline
Age (MAE) - Regression     & \textbf{\cite{savchenko2019efficient} 5.44 } & \textbf{6.21} & - \\ \hline
Age - Classification & \textbf{\cite{das2018mitigating} 70.1\%} &  - &   \textbf{67.94\%}  \\
\hline   
\end{tabular}
\end{table}

\subsection{Impact of Image Attributes}
\begin{figure}[htb]
    \centering
    \begin{subfigure}[b]{0.2\textwidth}
         \centering
         \includegraphics[width=\textwidth]{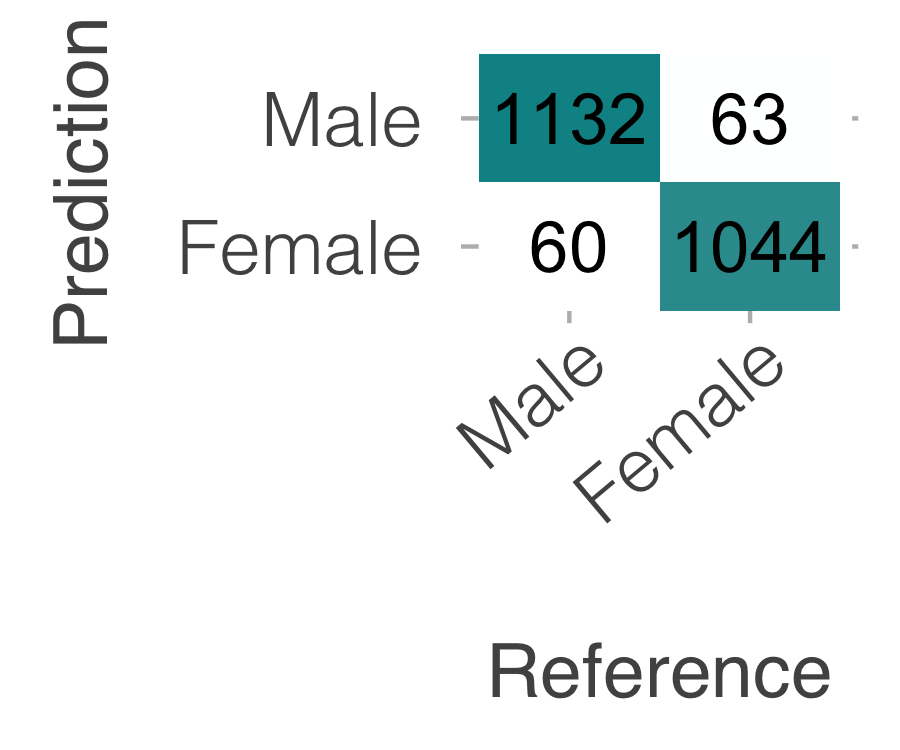}
         \caption{\textbf{Sex}}
         \label{fig:cf_sex}
     \end{subfigure}
     \hfill
     \begin{subfigure}[b]{0.33\textwidth}
         \centering
         \includegraphics[width=\textwidth]{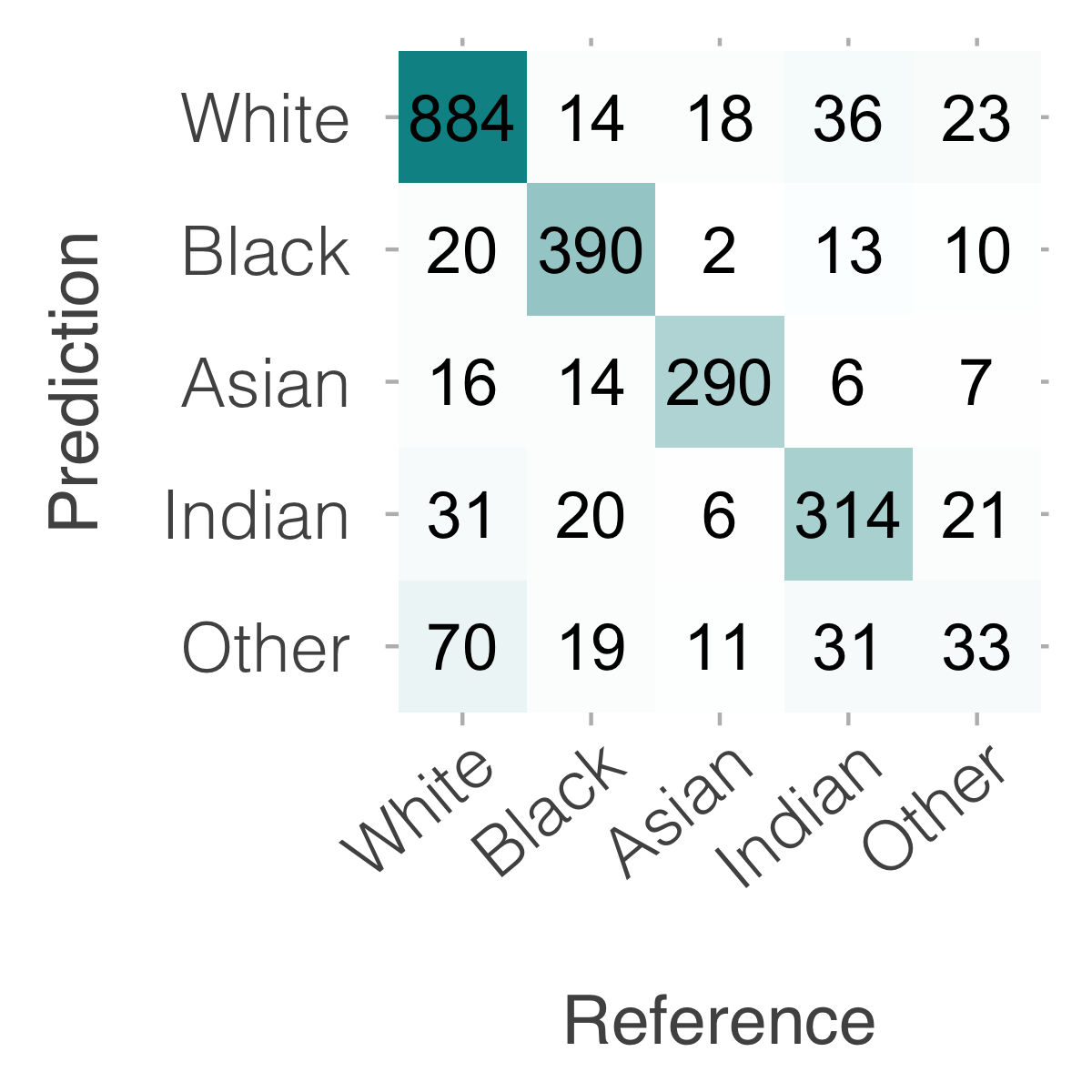}
         \caption{\textbf{Race}}
         \label{fig:cf_race}
     \end{subfigure}
     \hfill
     \begin{subfigure}[b]{0.43\textwidth}
         \centering
          \includegraphics[width=\textwidth]{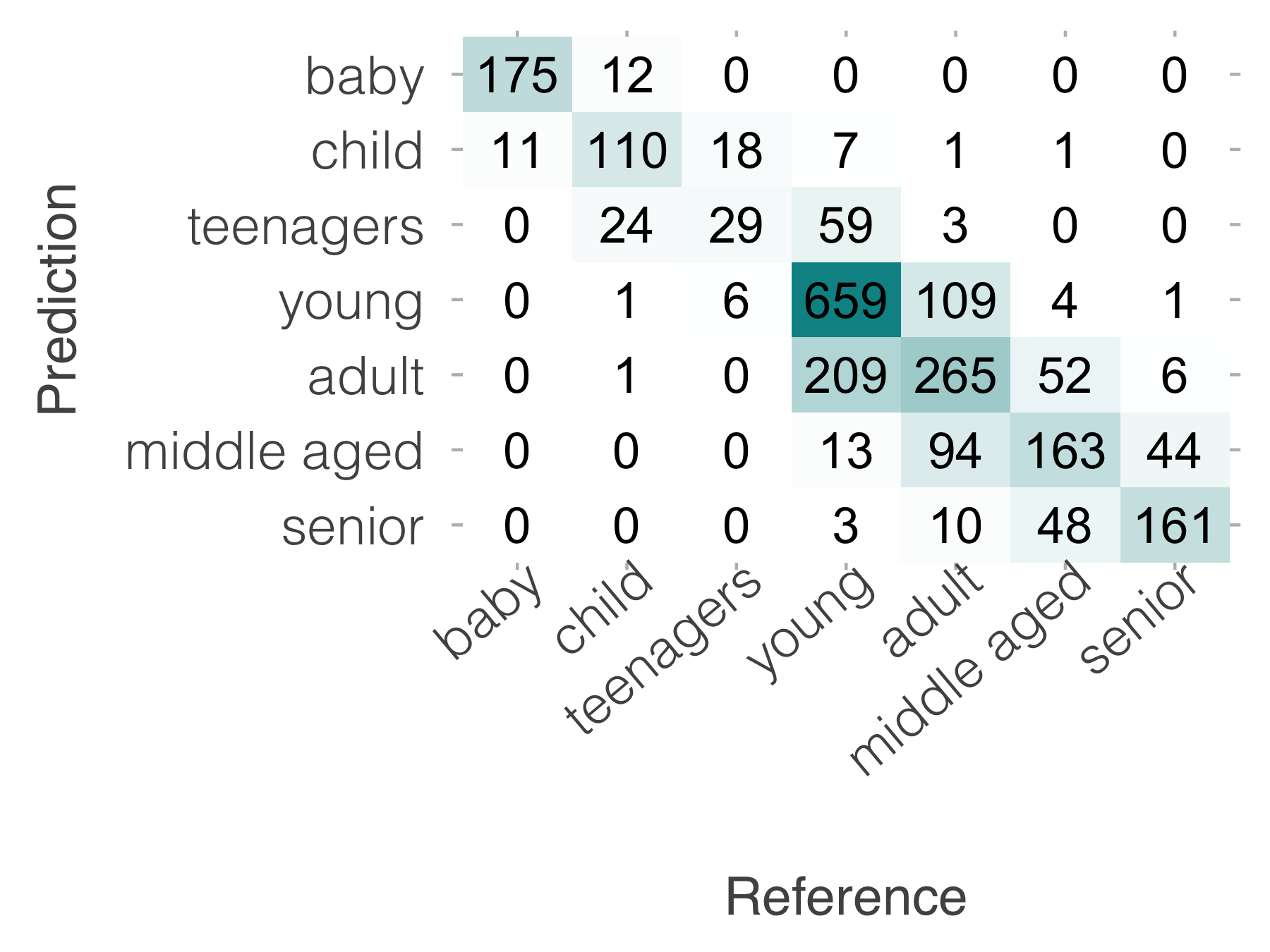}
         \caption{\textbf{Age}}
         \label{fig:cf_age}
     \end{subfigure}
    \caption{\textbf{Confusion matrices for Sex, Race and Age prediction using masked images.}}
    \label{fig:cfs}
\end{figure}

We examine whether the original user attributes (i.e., Sex, Race, Age category) influence our prediction outcome of masked faces. To this end, we select the best performing model from initial evaluation (Table~\ref{tab:model_results}) and evaluate sex, race and age prediction models using a new uniform test split with a balanced attribute composition. Fig.~\ref{fig:cfs} presents confusion matrices for each model outcome. We obtained an overall accuracy of 94.65\% for sex, 83.12\% for race and 67.94\% for age category.

\begin{figure}[htb]
    \centering
    \includegraphics[width=\textwidth]{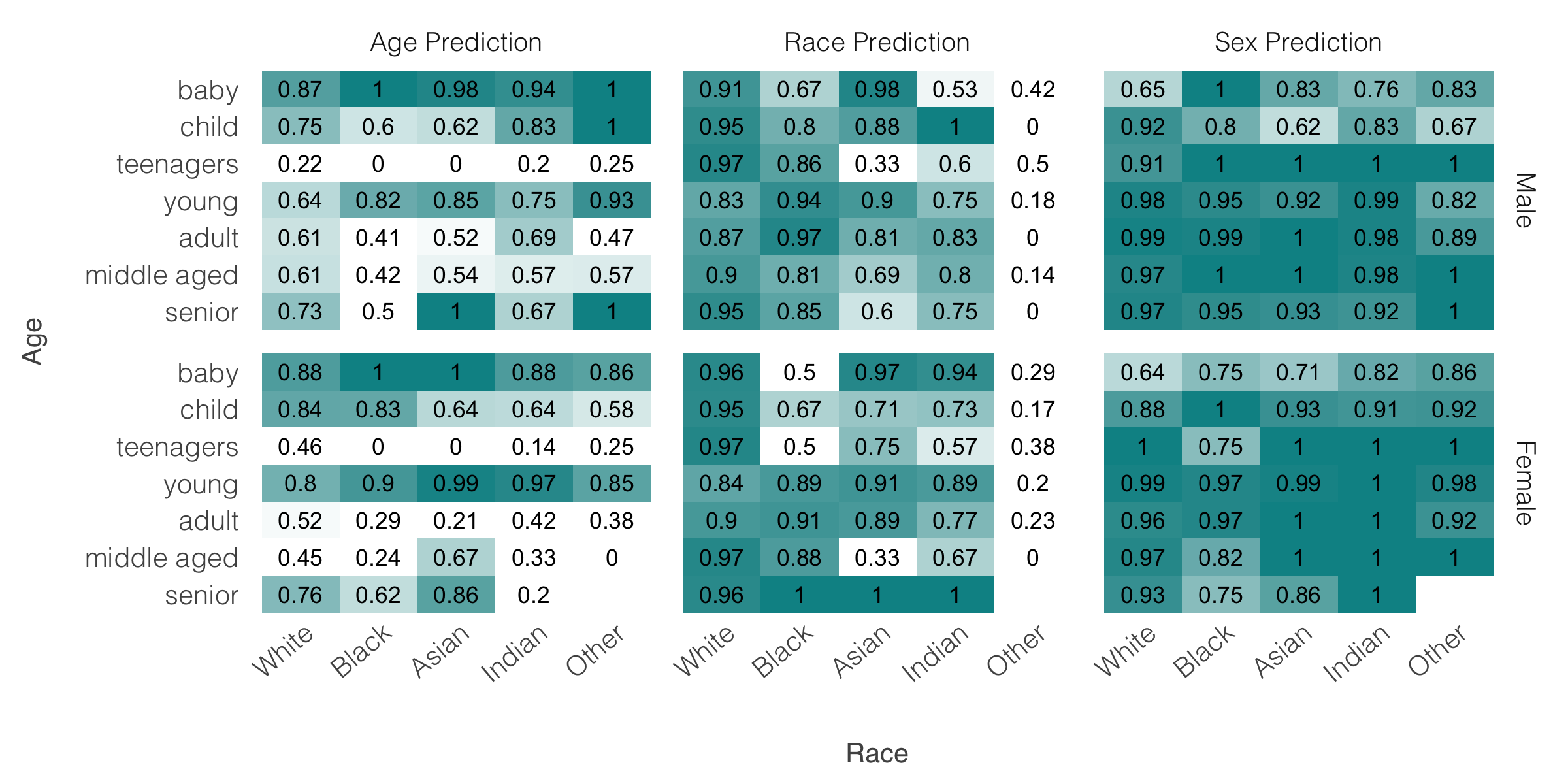}
    \caption{\textbf{Attribute prediction accuracy for each sub group}}
    \label{fig:subgroup_acc}
\end{figure}

We conduct chi-square tests of independence to examine the relationship between different image attributes and the ability to accurately predict the them. When considering the image attribute sex there is no significant difference between prediction outcomes of sex ($\chi^2$(1) = 0.006, p = 0.936) and race ($\chi^2$(1) = 0.578, p = 0.447). However, a significant difference is noted for age category ($\chi^2$(1) = 4.019, p = 0.045 $<$ 0.05). 
Furthermore, when considering race, there is a significant difference in prediction outcomes for sex ($\chi^2$(4) = 12.53, p = 0.014 $<$ 0.05), race ($\chi^2$(4) = 523.07, p $<$ 0.001 ) and age ($\chi^2$(4) = 49.951, p $<$ 0.001) prediction. 
Similarly, for image attribute age, there is a significant difference for outcomes of sex ($\chi^2$(6) = 164.57, p $<$ 0.001), race ($\chi^2$(6) = 13.449, 0.036 $<$ 0.05) and age ($\chi^2$(6) = 374.08, p $<$ 0.001) prediction. In summary, we note that image attributes race and age category having a significant impact on all the prediction outcomes while sex only influence age prediction. In addition, Fig.~\ref{fig:subgroup_acc} provides the accuracy for each subgroup of images based on sex, race, age category of the person appearing in the image.

\subsection{Privacy Invasiveness} 
We use the PVI equation with the SOTA for face images and our best results for masked face images to quantify the level of privacy invasiveness of both. \textbf{$PVI_f=\pviNoMask, PVI_{mf}=\pviMask$}. The privacy invasiveness reduction by wearing a mask is only \diffPvi\%. Which is very low compared to the 50\% of people who thought that masked faces to be more private.

\section{Discussion and Conclusion}
\subsection{Predicting Protected Attributes}

Our study shows it is possible to predict sex, race, and age with a high degree of accuracy. When compared to the state of the art methods that predict these attributes for non-masked face images, we only note absolute accuracy differences of 3.58\% for sex, 8.11\% for race and 2.16\% for age categories with nearly half the face (with key features like facial hair and lips for sex, wrinkles for age) occluded by a mask. Based on results in Table~\ref{tab:result_comparison}, we find that incorporating augmentations during training can improve sex and race prediction. During age prediction it slightly lowers accuracy. This likely stems from how predicting age is harder than race or sex (even for humans) and augmentations may create discrepancies between fine-grained features such as wrinkles which has less impact on predicting race or sex.




\subsection{Biases from Image Attributes}

While our models achieved high levels of overall accuracy for masked images, we observed that image attributes race and age can influence the prediction outcomes. For instance, age category prediction accuracy for teenagers (25.21\%) and adults (49.71\%) is low compared to the overall accuracy (67.94\%). As shown in Figure~\ref{fig:subgroup_acc}, prediction accuracies are consistently low across different sex and race categories as well. This is inline with results reported in prior work on biases in user attribute classification using regular face images~\cite{das2018mitigating}.
Furthermore, when considering race prediction, we note that race prediction accuracy is lower for Other category (20.12\%) with limited samples, when compared to the overall value (83.12\%). We argue that biases from user attributes can greatly influence the model outcomes. Therefore, appropriate measure should be taken to account for sampling biases particularly for commercial applications of face-recognition technology. 


\subsection{Privacy Preservation}

Our study highlights a mismatch between user perception and the reality regarding privacy preservation through face masks. Compared to regular face exposure, users perceive a significantly higher level of privacy when wearing face masks. However, we show that the ability to predict protected attributes from masked face images is not largely different from face images (Table~\ref{tab:result_comparison}) and the  privacy  invasiveness  reduction  by  wearing  a  mask  is  only  \diffPvi\%. This inaccurate perceived privacy could lead to a false sense of safety for masked users, and therefore users could be targets for exploitation by malicious applications. In addition, distinct characteristics of face masks could contribute to more robust surveillance applications that users are not aware of. In the light of increased make usage, we argue that it is essential to raise user awareness and research privacy protection methods concerning face masks.

\subsection{Limitations} 

We note a few limitations in our study. First, as there is no masked image dataset available with attributes such as sex, race and age, our evaluation is based on a synthetic mask generation process. Second, our user study is limited to 60 participants and we did not collect demographic information which may reveal interesting insights. Third, our source dataset has imbalances among classes which is reflected in our analysis.

\subsection{Conclusion and Future Work}


In this paper, we predict sex (\accSex\%), race (\accRace\%) and age (\accAge\%) on masked face images using a computer vision approach. Despite the popular belief that masks protect user privacy, we show that masks only reduce privacy invasiveness by \diffPvi\% when compared to state of the art face recognition approaches. We further analyse the impact of image labels on the prediction ability and provide a baseline for future research by open-sourcing our models. Our research paves the way for future work that
aim to study how to preserve user privacy when wearing masks while maintaining utility as
a biometric modality. We open-source our contributions, including masking and inference
scripts, trained models and data splits for reproducibility and broader use for both privacy
and mask related research.

\bibliographystyle{splncs04}
\bibliography{mendeley-ref,additional-ref}

\end{document}